\documentclass[11pt]{article}

\usepackage[preprint]{acl}

\usepackage{times}
\usepackage{latexsym}

\usepackage[T1]{fontenc}

\usepackage[utf8]{inputenc}

\usepackage{microtype}

\usepackage{inconsolata}

\usepackage{graphicx}
\usepackage{booktabs}
\usepackage{amsmath}
\usepackage{booktabs} 
\usepackage{multirow} 
\usepackage{makecell}
\usepackage[most]{tcolorbox}

\usepackage{float}
\usepackage{amssymb}
\usepackage{hyperref}

\title{AGGC: Adaptive Group Gradient Clipping for Stabilizing Large Language Model Training}

\author{
 \textbf{Zhiyuan Li\textsuperscript{1}},
 \textbf{Yuan Wu\textsuperscript{1}\thanks{Corresponding author},
 \textbf{Yi Chang\textsuperscript{1,2,3}}
 }
\\
 \textsuperscript{1} School of Artificial intelligence, JiLin University
 \\
 \textsuperscript{2} Engineering Research Center of Knowledge-Driven Human-Machine Intelligence, JiLin University
 \\
 \textsuperscript{3} International Center of Future Science, JiLin University
\\
 \small{
   \textbf{Correspondence:} \href{mailto:yuanwu@jlu.edu.cn}{yuanwu@jlu.edu.cn}
 }
}

\begin{document}
\maketitle

\begin{abstract}
To stabilize the training of Large Language Models (LLMs), gradient clipping is a nearly ubiquitous heuristic used to alleviate exploding gradients. However, traditional global norm clipping erroneously presupposes gradient homogeneity across different functional modules, leading to an adverse \textbf{"spill-over"} effect where volatile parameters force unnecessary scaling on stable ones. To overcome this, we propose Adaptive Group-wise Gradient Clipping (AGGC). AGGC partitions parameters into groups based on functional types and regulates each according to its historical behavior using an Exponential Moving Average (EMA). Specifically, it constructs an adaptive interval to simultaneously mitigate gradient explosion and vanishing, while employing a time-dependent scheduling mechanism to balance exploration and convergence. Experiments on LLaMA 2-7B, Mistral-7B, and Gemma-7B models show that AGGC consistently outperforms LoRA and frequently surpasses Full Fine-Tuning. On the GSM8K benchmark, Mistral-7B fine-tuned with AGGC achieves an accuracy of 72.93\%, exceeding LoRA’s 69.5\%. AGGC also effectively stabilizes Reinforcement Learning with Verifiable Rewards (RLVR), enhancing the logic deduction of Qwen 2.5 and Llama 3.2 models. Experimental results demonstrate that AGGC effectively addresses the limitations of traditional gradient clipping methods, particularly in overcoming gradient heterogeneity, by utilizing a modular, adaptive clipping strategy to stabilize the training process. Due to its lightweight design, AGGC can be seamlessly integrated into existing post-training pipelines with negligible overhead\footnote{Code is available at: \url{https://github.com/ZhiyuanLi218/AGGC}}.
\end{abstract}
\section{Introduction}
The relentless scaling of Large Language Models (LLMs) in terms of parameter count and depth has fundamentally reshaped the deep learning landscape, enabling unprecedented capabilities across diverse tasks~\citep{naveed2025comprehensive}. However, this scaling progression has simultaneously exacerbated foundational challenges in optimization, particularly concerning training stability. During both pre-training and post-training pipelines, such as Supervised Fine-Tuning (SFT) and Reinforcement Learning with Human Feedback (RLHF), LLMs frequently encounter catastrophic loss spikes~\citep{wang2025adagc,takase2023spike,ma2025understanding}. These spikes indicate transient or persistent numerical instability, which undermines training efficiency and compromises the quality of the resultant model convergence.

Gradient-based optimization necessitates a sophisticated trade-off between rapid convergence and numerical stability. As a canonical technique for safeguarding training stability, gradient clipping was originally developed to alleviate the critical issue of exploding gradients~\citep{pascanu2013difficulty}. By capping the magnitude of gradient updates, this method constrains the parameter updates of gradient descent within a rational range. Owing to its validated effectiveness as a regularization strategy, gradient clipping, particularly in the form of global norm clipping, has evolved into a nearly ubiquitous and indispensable heuristic for the training of contemporary deep neural networks~\citep{koloskova2023revisiting,liu2022communication,bu2023automatic,kenfack2022repfair}. \textbf{This assumption leads to the "spill-over" effect, where gradients from volatile modules with large magnitudes unnecessarily scale stable modules, disrupting the stability of the training process.}

Despite its ubiquity, conventional global gradient clipping bears inherent limitations rooted in transformer-based LLMs’ architectural traits. This standard method computes a unified $\ell_2$ norm across all parameters and erroneously presupposes gradient homogeneity—i.e., comparable magnitudes and variances across all functional modules~\citep{wu2025imbalanced,zhao2025helene,you2025gradient}.

\begin{figure}
    \centering
    \includegraphics[width=\linewidth]{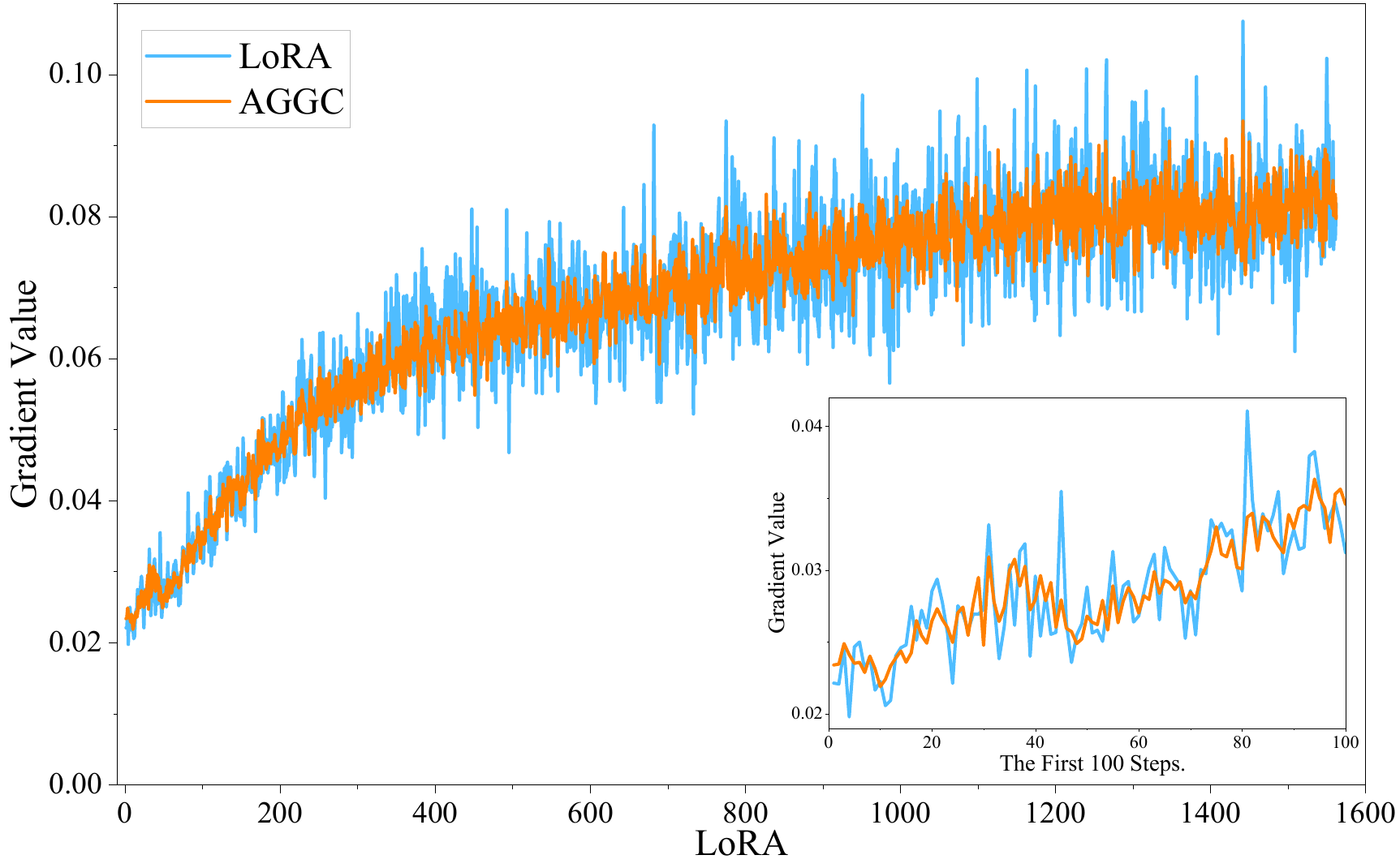}
    \caption{Grad norm of Up module over training steps.}
    \label{fig:grad_example}
\end{figure}

LLMs exhibit marked architectural heterogeneity: parameter groups (Query/Key/Value projections, Layer Normalization, Feed-Forward Networks) possess divergent gradient dynamics~\citep{wang2025adagc,tomihari2025understanding}. Empirically, gradient magnitudes vary drastically across components, with certain modules generating dominant large gradients~\citep{wu2025imbalanced}, rendering universal clipping inherently suboptimal—especially for models with heterogeneous loss curvatures~\citep{zhao2025helene}.

To address the inherent limitations of static global gradient clipping strategies, we propose a novel method termed Adaptive Group-wise Gradient Clipping (AGGC). The core design principle of AGGC lies in disentangling the optimization constraints across distinct functional components of the model.

Specifically, the gradient regulation mechanism for each component group is formulated as a dynamic process. AGGC leverages an Exponential Moving Average (EMA) to track the historical gradient norms of individual groups, generating a smoothed scaling statistic denoted as $S_j^{(t)}$, where $j$ represents the group and $t$ denotes the training step. 

Furthermore, AGGC introduces a bidirectional admissible interval defined by $[\mathrm{L}_j^{(t)},\mathrm{U}_j^{(t)}]$ for each group $j$ at training step $t$. The upper bound $\mathrm{U}_j^{(t)}$ is designed to suppress gradient explosion, while the lower bound $\mathrm{L}_j^{(t)}$ proactively alleviates the risks of gradient vanishing and premature stagnation of parameter updates. 
Critically, the width of this interval is modulated by a time-dependent scheduling strategy. Specifically, the bounds \( \mathrm{L}_j^{(t)} \) and \( \mathrm{U}_j^{(t)} \) are controlled by the multiplicative coefficients \( \alpha_{\mathrm{low}}^{(t)} \) and \( \alpha_{\mathrm{high}}^{(t)} \), which define the lower and upper limits of the gradient norm at each training step \( t \). In the early training phase, these coefficients are set relatively high to allow for sufficient parameter exploration, and they gradually decrease to stabilize the model's convergence in later stages.
As illustrated in Figure~\ref{fig:grad_example}, AGGC maintains a smoother and more stable optimization trajectory compared to the fluctuations observed in LoRA.

The AGGC framework contributes significantly to the field of optimization by providing a robust, nuanced control mechanism tailored for LLMs:
\begin{enumerate}
    \item \textbf{Functional Grouping and Localized Regulation:} We formalize a \textit{Group-wise} gradient regulation strategy, partitioning parameters based on functional module type to align optimization control with architectural heterogeneity. This design successfully eliminates the adverse spill-over effect inherent in global clipping, ensuring efficient and unbiased utilization of small-scale gradient signals.
    \item \textbf{Adaptive and Bi-directional Control:} AGGC leverages EMA to dynamically estimate group-specific gradient scales, establishing an adaptive, two-sided admissible interval $[\mathrm{L}_j^{(t)}, \mathrm{U}_j^{(t)}]$. This approach actively mitigates both exploding gradients and update collapse, a capability largely absent in standard clipping methods.
    \item \textbf{Integration of Magnitude Scheduling:} We introduce a time-dependent, linear scheduling mechanism for the multiplicative coefficients ($\alpha_{\mathrm{low}}^{(t)}, \alpha_{\mathrm{high}}^{(t)}$) that define the bounds. This provides a principled, dedicated method for gradient magnitude control scheduling, optimally balancing exploratory dynamics early in training with stable convergence in later stages.
\end{enumerate}

\section{Related Work}
The theoretical foundation of traditional gradient clipping originated from the in-depth analysis of training difficulties in recurrent neural networks (RNNs). \citep{pascanu2013difficulty} presented a seminal work that systematically explored gradient vanishing and explosion issues from analytical, geometric, and dynamic system perspectives. They proposed gradient norm clipping to mitigate exploding gradients and soft constraint methods for vanishing gradients, laying the theoretical groundwork for all subsequent gradient clipping approaches. Global norm clipping's core idea is to limit the maximum magnitude of gradients using a unified global threshold. This effectively prevents numerical instability and gradient explosion~\citep{koloskova2023revisiting}.

Nevertheless, traditional global clipping exhibits inherent limitations. \citep{koloskova2023revisiting} pointed out in their comprehensive review that despite its simplicity and widespread adoption, gradient clipping mechanisms typically require specific threshold values \( c \) and strong noise assumptions to guarantee convergence. Another pivotal observation came from \citep{zhang2019gradient}, who found that gradient smoothness exhibits significant variability along the training trajectory of deep neural networks, and this smoothness is positively correlated with gradient norm. To overcome the limitations of traditional global clipping, researchers have developed a series of adaptive clipping techniques that dynamically adjust strategies based on gradient characteristics during training.

Early adaptive methods such as AdaGrad~\citep{duchi2011adaptive} and RMSProp indirectly addressed gradient adaptation by tracking historical gradient information to adjust learning rates dynamically. However, their focus remained on learning rate adaptation rather than direct gradient clipping optimization.  Adaptive Gradient Clipping (AdagC), proposed by~\citep{wang2025adagc}, represents the latest advancement in adaptive clipping technology. The AdagC framework dynamically adjusts local clipping thresholds for each parameter using an Exponential Moving Average (EMA) mechanism. 

Recent advances in gradient clipping have increasingly focused on module-based gradient regulation, which tailors optimization strategies to the heterogeneous structures of LLMs~\citep{zhao2025helene}. Prior work has identified gradient imbalance as a critical bottleneck, particularly in post-training multi-task reinforcement learning. \citep{wu2025imbalanced} showed that gradients from different tasks can differ by up to 15--33$\times$, causing optimization to disproportionately favor large-gradient tasks while neglecting others, and further revealed that gradient magnitude does not reliably reflect task-wise learning progress. To mitigate such issues, several module-level methods have been proposed. AlphaDecay~\citep{he2025alphadecay} proposes module-level adaptive weight decay based on the heavy-tailedness of each module’s empirical spectral density, aligning regularization strength with functional importance. Beyond single-task optimization, module-based regulation has also been explored in multi-task learning through gradient clipping techniques. \citep{yu2020gradient} proposed resolving task conflicts by projecting gradients onto the normal plane of interfering gradients, demonstrating consistent efficiency and performance gains across supervised and reinforcement learning settings. Collectively, these works highlight the effectiveness of module gradient control as a unifying strategy for addressing gradient imbalance and architectural heterogeneity.

Despite these promising advancements in adaptive and module-aware optimization algorithms, a critical research gap persists in the simultaneous mitigation of architectural heterogeneity and the bidirectional characteristics of training instability. Most existing strategies either adopt rigid global constraints, which consequently induce the adverse gradient spill-over effect, or focus exclusively on alleviating gradient explosion while leaving the risks of gradient vanishing and premature training stagnation largely unaddressed. Furthermore, few methods explicitly modulate the gradient clipping intensity to achieve an optimal trade-off between the conflicting objectives of parameter space exploration during the early phase of training and precise convergence during the later phase. To address these aforementioned challenges, we propose a novel method termed AGGC. By leveraging exponential moving averages to construct dynamic and group-specific gradient admissible intervals, AGGC decouples the optimization constraints across heterogeneous functional components and integrates a time-dependent scheduling mechanism to maintain training stability throughout the entire lifecycle of model training.
\section{Method}

\begin{figure}
    \centering
    \includegraphics[width=\linewidth]{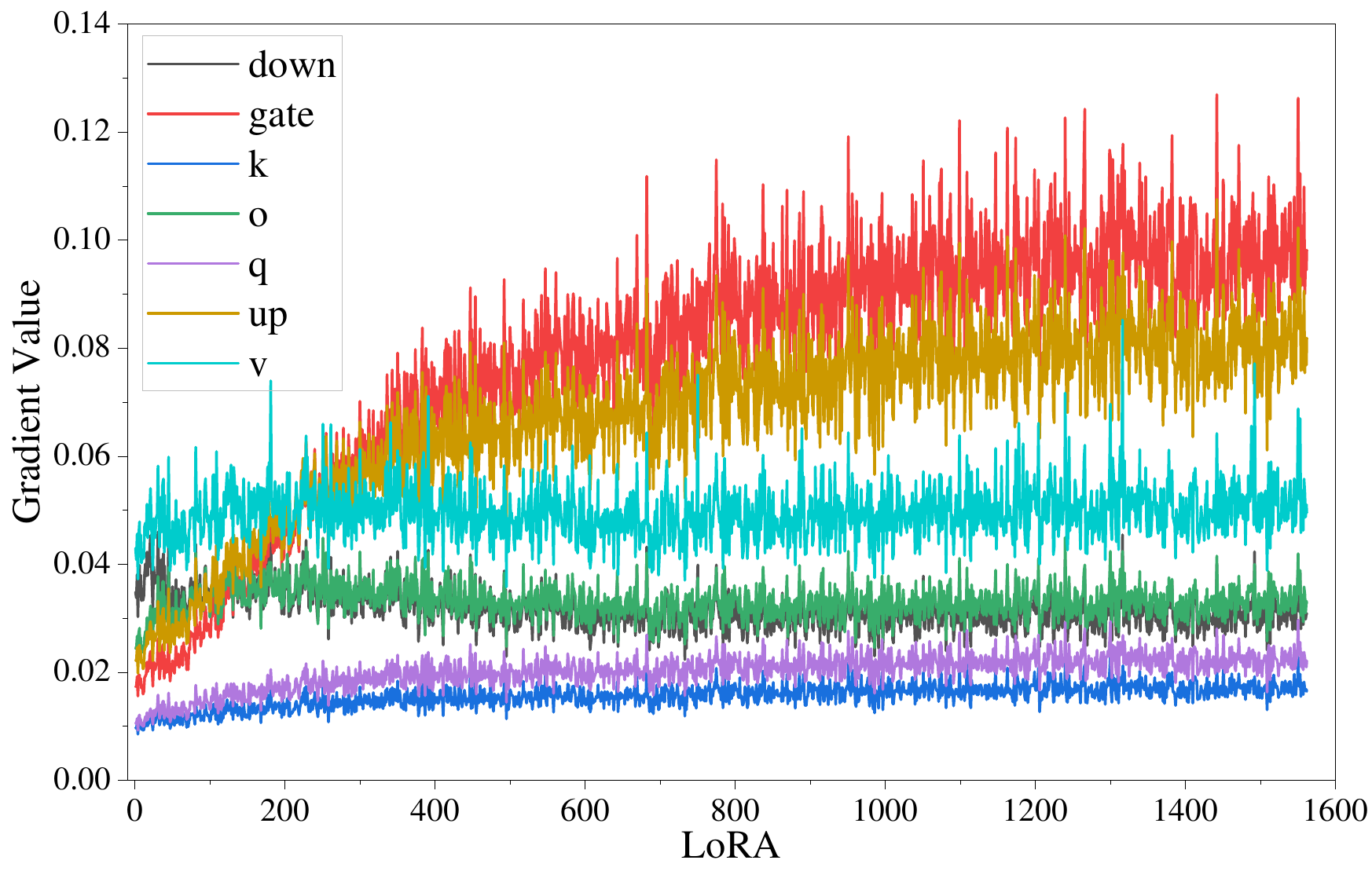}
    \caption{Gradient norm evolution across parameter groups during training.}
    \label{fig:motivation}
\end{figure}

\begin{figure*}
    \centering
    \includegraphics[width=\linewidth]{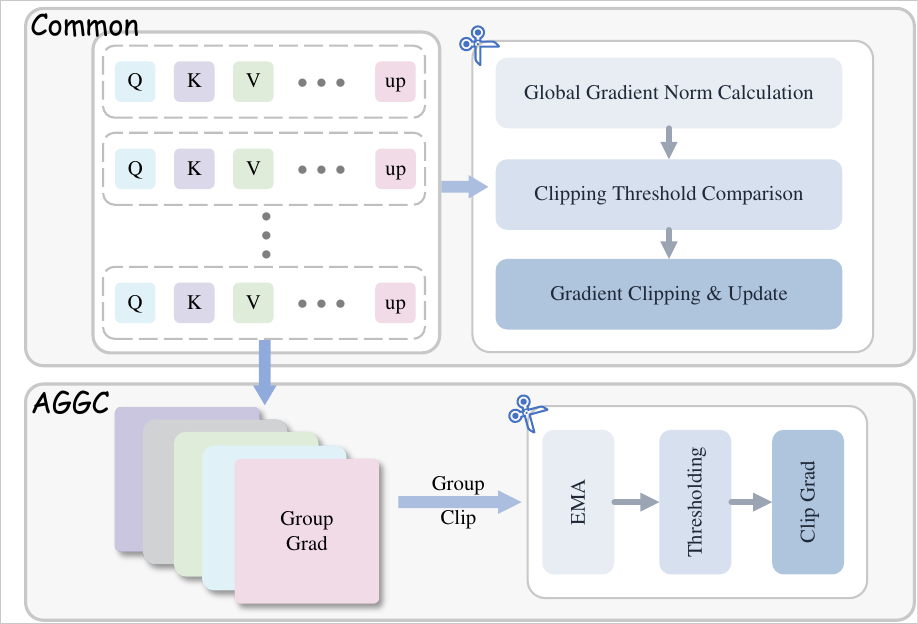}
    \caption{The comparison between AGGC and conventional gradient adjustment methods.}
    \label{fig:main}
\end{figure*}

Gradient clipping in current LLM post-training pipelines typically relies on a global norm computed over all parameters. While simple, this strategy implicitly assumes that gradients across different components of the model evolve with comparable magnitudes. As illustrated in Figure~\ref{fig:motivation}, we observe that the gradients of certain parameters (i.e., the gate, up, and value) evolve smoothly and stay within a relatively narrow range, whereas those of other parameters may undergo significant fluctuations with large amplitudes during training. This mismatch leads to unstable optimization dynamics and inefficient gradient utilization. As shown in the Figure~\ref{fig:main}, to address this discrepancy, we introduce AGGC. The central idea is to aggregate all parameters belonging to the same type of module into a single gradient group, and to regulate each group according to its own historical gradient behavior.

Let the model parameters be partitioned into $J$ groups $\{G_j\}_{j=1}^J$, where each group corresponds to a same module (for example, all query vectors or all value vectors). At training step $t$, let the gradients in group $G_j$ be $\{g_{j,i}^{(t)}\}_{i=1}^{N_j}$. We define the group gradient norm as
\begin{equation}
\|\nabla_{G_j}^{(t)}\|_2 = \left( \sum_{i=1}^{N_j} \| g_{j,i}^{(t)} \|_2^2 \right)^{1/2},
\label{eq:instantaneous_group_norm}
\end{equation}
where $N_j$ denotes the number of parameter tensors in group $G_j$, and $\|\nabla_{G_j}^{(t)}\|_2$ measures the $L_2$ magnitude of the group's aggregated gradient at step $t$.

To model the temporal trend of each group's gradient magnitude, we estimate its scale at step $t$ by applying an exponential moving average (EMA) to its historical gradient norms, producing a smoothed statistic that reflects the group’s long-term behavior. Concretely, we maintain
\begin{equation}
S_j^{(t)} \;=\; \beta\, S_j^{(t-1)} + (1-\beta)\,\|\nabla_{G_j}^{(t)}\|_2,
\label{eq:ema_corrected}
\end{equation}
where $S_j^{(t)}$ denotes the EMA-based scale of group $G_j$ at step $t$, $\beta\in[0,1)$ is the decay factor controlling the estimator's memory.

Based on the EMA \(S_j^{(t)}\) we construct an adaptive admissible interval \([\mathrm{L}_j^{(t)},\mathrm{U}_j^{(t)}]\) for the group's gradient norm. The lower bound is given by
\begin{equation}
\mathrm{L}_j^{(t)} \;=\; \max\!\big(\mathrm{min\_norm},\; \alpha_{\mathrm{low}}^{(t)}\, S_j^{(t)}\big),
\label{eq:lower_bound}
\end{equation}
and the upper bound by
\begin{equation}
\mathrm{U}_j^{(t)} \;=\; \alpha_{\mathrm{high}}^{(t)}\, S_j^{(t)}.
\label{eq:upper_bound}
\end{equation}
In these expressions \(\mathrm{min\_norm}\ge 0\) prevents the lower bound from collapsing to zero, and \(\alpha_{\mathrm{low}}^{(t)}\) and \(\alpha_{\mathrm{high}}^{(t)}\) are multiplicative coefficients that determine the tolerated deviation from the gradient. The coefficients are time-dependent in order to allow more permissive behavior early in training and tighter control during convergence.

The multiplicative coefficients governing the admissible interval play a critical role in shaping the optimization trajectory. If the interval is excessively restrictive at early training stages, gradients are subjected to frequent rescaling, which can induce premature contraction of the update magnitudes, accelerate entropy decay, and inhibit adequate exploration of the parameter space~\citep{pascanu2013difficulty,wang2025adagc,kim2024knowledge}. Such behavior increases the likelihood of convergence toward suboptimal regions and may adversely affect final model performance. To mitigate this effect, and drawing an analogy to the rationale underlying learning-rate schedules, we employ a time-dependent adjustment in which the interval is broader at the beginning of training and progressively tightened as optimization proceeds. This design facilitates exploratory dynamics initially while promoting stability and controlled convergence in later stages.

Formally, we implement this progression by applying a linear schedule to each coefficient \(\alpha^{(t)}\):
\begin{equation}
\alpha^{(t)} = 
\begin{cases}
    \alpha_{\mathrm{init}}, & p \le s, \\[2pt]
    (1-\lambda)\alpha_{\mathrm{init}} + \lambda\alpha_{\mathrm{late}}, & s < p < s+w, \\[2pt]
    \alpha_{\mathrm{late}}, & p \ge s+w,
\end{cases}
\end{equation}
\[
\lambda = \dfrac{p-s}{w},\qquad p = \frac{t}{T}.
\]
where \(T\) is the total number of optimization steps, \(s\) denotes the onset of the transition, \(w\) defines the duration of the transition window, and \(\alpha^{\mathrm{init}}\) and \(\alpha^{\mathrm{late}}\) correspond to the initial and final coefficient values.

Having established the adaptive interval \([\mathrm{L}_j^{(t)},\mathrm{U}_j^{(t)}]\) for group \(G_j\), we determine whether the group norm $\|\nabla_{G_j}^{(t)}\|_2$ lies inside this interval. If \(\|\nabla_{G_j}^{(t)}\|_2\) falls within the interval, no modification is performed; otherwise the group's gradients are multiplicatively rescaled so that the post-scaling norm attains the nearest interval boundary. Formally, with \(\varepsilon>0\) a small numerical constant, the unclipped scaling factor is computed as
\begin{equation}
c_j^{(t)} \;=\;
\begin{cases}
\dfrac{\mathrm{U}_j^{(t)}}{\|\nabla_{G_j}^{(t)}\|_2 + \varepsilon}, & \text{if }\|\nabla_{G_j}^{(t)}\|_2 > \mathrm{U}_j^{(t)}, \\[10pt]
\dfrac{\mathrm{L}_j^{(t)}}{\|\nabla_{G_j}^{(t)}\|_2 + \varepsilon}, & \text{if }\|\nabla_{G_j}^{(t)}\|_2 < \mathrm{L}_j^{(t)}, \\[6pt]
1, & \text{otherwise}.
\end{cases}
\label{eq:raw_scaling_rewritten}
\end{equation}

The group-wise factor is applied uniformly across the group's gradients:
\begin{equation}
g_{j,i}^{(t)} \leftarrow c_j^{(t)}\, g_{j,i}^{(t)}, \qquad \forall i=1,\dots,N_j.
\label{eq:apply_scaling_rewritten}
\end{equation}

In summary, AGGC provides a lightweight mechanism to regulate gradient magnitudes at the module-group level. By estimating group-specific scales via EMA, constructing adaptive admissible intervals, and applying bounded multiplicative rescaling only when necessary. AGGC mitigates adverse spill-over from high-amplitude gradients, and preserves useful small-scale signals. The method imposes minimal computational and memory overhead and integrates transparently with existing optimization pipelines used in LLM post-training. For further details on GPU memory usage, please refer to Appendix~\ref{appendix:resource}.

\section{Experiment}
The experiments were conducted on four NVIDIA A40 GPU. In our experiments, for LoRA fine-tuning~\citep{hu2022lora}, we employed the AdamW optimizer with a learning rate of 2E-5, cosine annealing with a warm-up rate of 0.03, and no weight decay. We ensured lora\_alpha always equaled lora\_r, set lora\_dropout to 0, and merged the adapter into all linear layers of the base model. We employed the Float32 computation type for both the base model and the adapters within LoRA and AGGC. For the GRPO experiment~\citep{shao2024deepseekmath}, the learning rate was set to 1e-6, weight decay was 0.01, and the batch size was set to 512. Detailed experimental parameter settings are provided in the appendix~\ref{appendix:param_nlg}.

\begin{table*}[htbp]
  \centering
    \begin{tabular}{ccccccc}
    \toprule
    Model & Strategy & GSM8K & MATH  & HumanEval & MBPP  & MT-Bench \\
    \midrule
    \multirow{3}[2]{*}{LLaMA 2-7B} & Full FT & \textbf{49.13} & \textbf{7.29} & 21.2  & 35.59 & \textbf{4.91} \\
          & LoRA  & 42.85 & 5.5   & 18.35 & 35.5  & 4.59 \\
          & AGGC  & 48.06 & 6.64  & \textbf{21.3} & \textbf{37.8} & 2.95 \\
    \midrule
    \multirow{3}[2]{*}{Mistral-7B} & Full FT & 69.91 & 18.64 & 45.31 & 51.46 & 4.95 \\
          & LoRA  & 69.5  & 20.08 & 43.78 & 58.46 & 4.9 \\
          & AGGC  & \textbf{72.93} & \textbf{21.42} & \textbf{47.6} & \textbf{65.1} & \textbf{5.86} \\
    \midrule
    \multirow{3}[2]{*}{Gemma-7B} & Full FT & 72.09 & 22.71 & 47.02 & 55.67 & 5.4 \\
          & LoRA  & 75.11 & \textbf{30.41} & 53.7  & 65.58 & 4.98 \\
          & AGGC  & \textbf{78.54} & 29.84 & \textbf{54.3} & \textbf{66.4} & \textbf{6.54} \\
    \bottomrule
    \end{tabular}%
  \caption{Experimental results on NLG tasks.}
  \label{tab:nlg}%
\end{table*}%

We evaluate the natural language generation capabilities of LLaMA 2-7B~\citep{touvron2023llama}, Mistral-7B~\citep{jiang2023mistral7b}, and Gemma-7B~\citep{team2024gemma} under a AGGC-enhanced LoRA framework through mathematical reasoning, code generation, and dialogue tasks. In addition, we assess natural language understanding performance using the GLUE benchmark~\citep{wang2018glue} in conjunction with the DeBERTa-v3-base model~\citep{he2021debertav3}. Furthermore, we investigate the effectiveness of applying AGGC to the GRPO framework, conducting experiments on the Math~\citep{yu2023metamath} and GSM8K~\citep{cobbe2021training} datasets with Qwen2.5-3B Instruct, Qwen2.5-1.5B Instruct~\citep{qwen2025qwen25technicalreport}, and LLaMA 3.2-3B Instruct models~\citep{dubey2024llama}.

\subsection{Experiments on Natural Language Generation}
Our experimental studies were conducted using the LLaMA 2-7B, Mistral-7B-v0.1 and Gemma-7B models. To evaluate mathematical reasoning capabilities, the models were fine-tuned on the MetaMathQA dataset and subsequently assessed on the GSM8K and MATH benchmarks. For the evaluation of coding proficiency, the models were fine-tuned using the CodeFeedback dataset~\citep{zheng2024opencodeinterpreter}, with performance measured on the HumanEval~\citep{chen2021evaluating} and MBPP~\citep{austin2021program} benchmark suites. To assess conversational abilities, the models were fine-tuned on the WizardLM-Evol-Instruct dataset~\citep{xu2024wizardlm} and evaluated using the MT-Bench benchmark~\citep{zheng2023judging}. All experiments were conducted on a 100K-sample subset of the corresponding datasets.

As presented in Table~\ref{tab:nlg}, the proposed AGGC strategy demonstrates consistent improvements over the standard LoRA baseline across the evaluated models, particularly for Mistral-7B and Gemma-7B where it frequently surpasses Full Fine-Tuning (Full FT) in mathematical reasoning and code generation tasks.  It is worth noting that the anomalous performance drop observed for LLaMA 2-7B on MT-Bench is not indicative of reduced reasoning capability, but rather stems from the "failure to stop" generation issue (i.e., the model failing to emit termination tokens), a known instability phenomenon documented in prior studies~\citep{yao2025understanding} and further analyzed in Appendix~\ref{appendix:failure_to_stop}.  
In addition, to further examine the robustness of AGGC under different adapter capacities, we conduct a systematic analysis across a wide range of LoRA ranks. Detailed experimental results and discussions are provided in Appendix~\ref{appendix:rank}.

\begin{table*}[htbp]
  \centering
    \begin{tabular}{cccccccccc}
    \toprule
    Method & MNLI  & SST2  & MRPC  & CoLA  & QNLI  & QQP   & RTE   & STSB  & ALL \\
    \midrule
    Full FT & 89.9  & 95.63 & 89.46 & 69.19 & 94.03 & \textbf{92.4} & 83.75 & 91.6  & 88.25 \\
    BitFit & 89.37 & 94.84 & 87.75 & 66.96 & 92.24 & 88.41 & 78.7  & 91.35 & 86.2 \\
    HAdapter & 90.13 & 95.53 & 89.95 & 68.64 & 94.11 & 91.91 & 84.48 & 91.48 & 88.28 \\
    PAdapter & 90.33 & 95.61 & 89.46 & 68.77 & 94.29 & 92.04 & 85.2  & 91.54 & 88.41 \\
    LoRA  & \textbf{90.65} & 94.95 & 89.95 & 69.82 & 93.87 & 91.99 & 85.2  & 91.6  & 88.5 \\
    DoRA  & 90.29 & 95.79 & 90.93 & \textbf{70.85} & 94.1  & 92.07 & 86.04 & 91.79 & 88.98 \\
    AGGC  & 90.59 & \textbf{96.33} & \textbf{93.03} & 70.8  & \textbf{94.49} & 92.32 & \textbf{90.98} & \textbf{92.29} & \textbf{90.1} \\
    \bottomrule
    \end{tabular}%
  \caption{Experimental results on NLU tasks.}
  \label{tab:nlu}%
\end{table*}%

\subsection{Experiments on Natural Language Understanding}
We further evaluated the Natural Language Understanding (NLU) capabilities of DeBERTa-v3 based on the GLUE benchmark. Table~\ref{tab:nlu} summarizes the results of the eight tasks included in the GLUE benchmark.

As illustrated in Table~\ref{tab:nlu}, our proposed AGGC strategy demonstrates superior performance across the GLUE benchmark compared to both Full FT and various parameter-efficient fine-tuning (PEFT) methods. AGGC achieves the highest accuracy in the majority of individual tasks and secures the best overall average score among all evaluated strategies. Notably, in challenging tasks requiring complex linguistic inference, such as RTE and MRPC, AGGC significantly outperforms the standard LoRA baseline. While modern techniques like DoRA~\citep{liu2024dora} offer incremental gains over standard adapters, AGGC consistently provides a more substantial performance boost by effectively addressing gradient heterogeneity. These results verify that our adaptive group-wise clipping mechanism is not only effective for generative tasks but also serves as a powerful optimizer for enhancing the discriminative capabilities of language models.

\subsection{Ablation Study}
\subsubsection{Effectiveness of Time-Varying Bounds}
We conduct an ablation study to evaluate the effectiveness of the proposed time-varying lower and upper bound coefficients. As shown in Table~\ref{tab:ablation_bounds}, enabling the scheduled bounds consistently improves performance over the static variant across both models and benchmarks. These consistent improvements indicate that dynamically adjusting the admissible gradient interval over training better balances early-stage flexibility and late-stage stability, thereby leading to more effective optimization than fixed bounds.

\begin{table}[htbp]
  \centering
    \begin{tabular}{cccc}
    \toprule
    Model & Strategy & GSM8K & MATH \\
    \midrule
    \multirow{2}[2]{*}{Mistral-7B} & $\times$     & 71.39 & 21.02 \\
          & \checkmark    & \textbf{72.93} & \textbf{21.42} \\
    \midrule
    \multirow{2}[2]{*}{Gemma-7B} & $\times$     & 76.34 & 29.56 \\
          & \checkmark      & \textbf{78.54} & \textbf{29.84} \\
    \bottomrule
    \end{tabular}%
  \caption{Ablation study on time-varying bound coefficients.}
  \label{tab:ablation_bounds}%
\end{table}%

\subsubsection{Effect of the EMA Decay Factor $\beta$}
We further investigate the effect of the EMA decay factor $\beta$ on mathematical reasoning performance. As shown in Table~\ref{tab:ablation_beta}, the MATH accuracy of Mistral-7B exhibits a consistent upward trend as $\beta$ increases from 0.1 to 0.95, reaching its highest value at $\beta=0.95$. This behavior suggests that a larger $\beta$ leads to a smoother and more stable estimation of the group-wise gradient scale $S^{(t)}_{j}$, which in turn yields more reliable admissible intervals and reduces noise in the gradient clipping process.

\begin{table}[htbp]
  \centering
    \begin{tabular}{ccc}
    \toprule
    Model & Beta  & MATH \\
    \midrule
    \multirow{5}[2]{*}{Mistral-7B} & 0.1   & 20.76 \\
          & 0.5   & 21.22 \\
          & 0.7   & 21.3 \\
          & 0.9   & 21.22 \\
          & 0.95  & \textbf{21.42} \\
    \bottomrule
    \end{tabular}%
  \caption{Ablation Study on the EMA Decay Factor $\beta$ on MATH.}
  \label{tab:ablation_beta}%
\end{table}%

\begin{figure*}
    \centering
    \includegraphics[width=\linewidth]{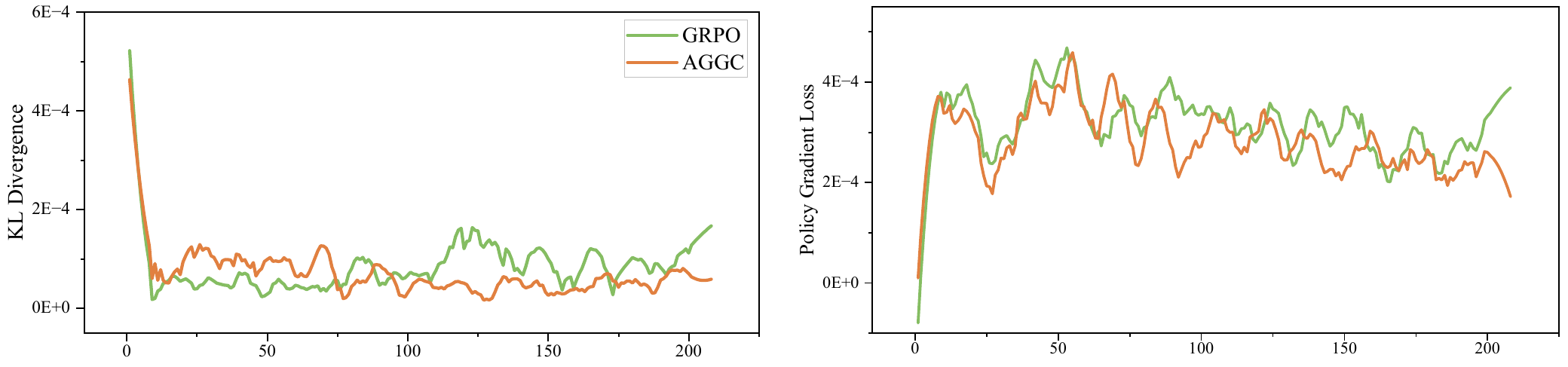}
    \caption{Comparison of KL divergence and PG loss in GRPO training.}
    \label{fig:grpo_index}
\end{figure*}

\subsection{Experiments on RLVR}
The effectiveness of our AGGC strategy was further evaluated within the GRPO framework across several instruction-tuned models. As shown in Table~\ref{tab:rl}, AGGC consistently outperforms both the base models and the standard GRPO training across mathematical reasoning benchmarks. 

\begin{table}[htbp]
  \centering
  \resizebox{\linewidth}{!}{
    \begin{tabular}{cccc}
    \toprule
    Model & Strategy & GSM8K & MATH \\
    \midrule
    \multirow{3}[2]{*}{Qwen 2.5 3B Instruct} & Base  & 86.3  & 66.1 \\
          & GRPO  & \textbf{87.9} & 66.9 \\
          & AGGC  & \textbf{87.9} & \textbf{67.2} \\
    \midrule
    \multirow{3}[2]{*}{Qwen 2.5 1.5B Instruct} & Base  & 73.8  & 55.5 \\
          & GRPO  & 77.6  & 58.6 \\
          & AGGC  & \textbf{79.8} & \textbf{59.1} \\
    \midrule
    \multirow{3}[2]{*}{Llama 3.2 3B Instruct} & Base  & 78.5  & 47.2 \\
          & GRPO  & 81.4  & 49 \\
          & AGGC  & \textbf{82.3} & \textbf{50.2} \\
    \bottomrule
    \end{tabular}%
  }
  \caption{Experimental results on RLVR.}
  \label{tab:rl}%
\end{table}%

\begin{figure}
    \centering
    \includegraphics[width=\linewidth]{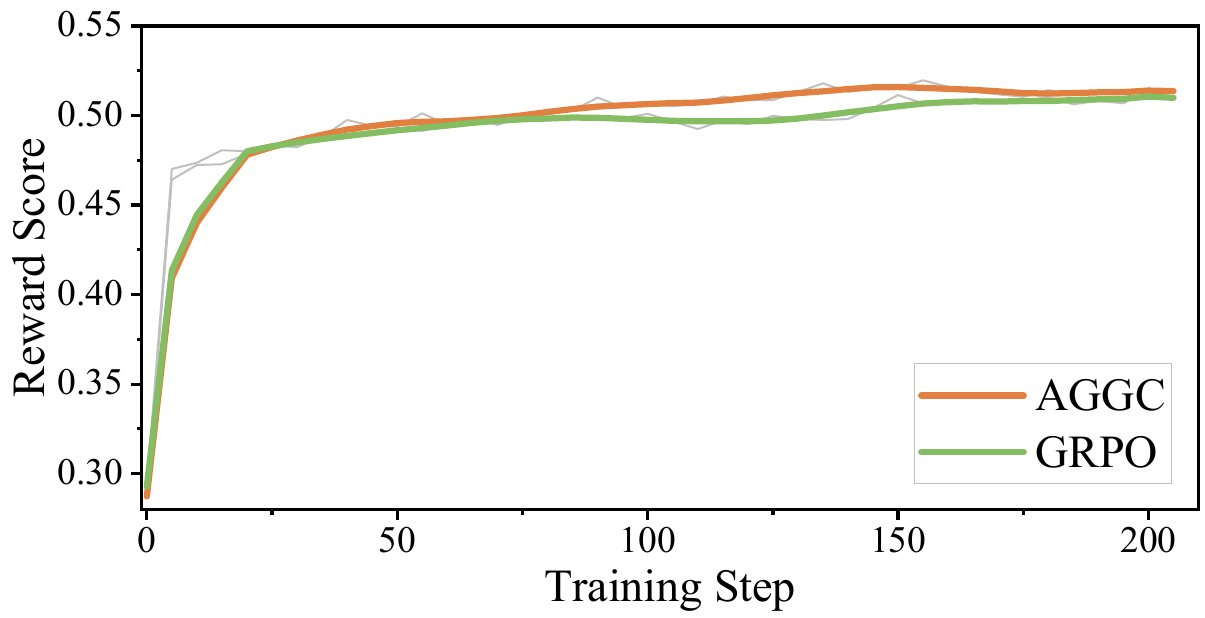}
    \caption{Evolution of Reward Scores during the Training Process of GRPO and AGGC.}
    \label{fig:reward_socre}
\end{figure}

To further elucidate the training dynamics and stability, we analyze the evolution of key metrics. As illustrated in Figure~\ref{fig:reward_socre}, the reward score of AGGC consistently exceeds that of the LoRA baseline throughout the training process, indicating that our method facilitates more effective policy improvement. This is complemented by the training trajectories shown in Figure~\ref{fig:grpo_index}, which compares the KL divergence and Policy Gradient (PG) loss. Specifically, AGGC exhibits a higher KL divergence than standard GRPO during the initial training stages, suggesting enhanced exploration. In the later stages, however, the KL divergence of AGGC drops below that of GRPO, signifying a more stable convergence. Simultaneously, the PG loss for AGGC remains lower than that of GRPO throughout the entire training duration. These trends collectively demonstrate that AGGC significantly reduces optimization variance and provides a smoother training trajectory.

Specifically, when applied to the Qwen 2.5 and Llama 3.2 series, AGGC demonstrates a superior ability to enhance the models' logical deduction capabilities compared to the original GRPO. In smaller scale models like Qwen 2.5 1.5B, the advantage of AGGC is particularly evident, achieving the highest accuracy on both GSM8K and MATH datasets. These results indicate that by regulating gradient magnitudes through group-wise adaptive clipping, AGGC effectively stabilizes the reinforcement learning process, which is often prone to high variance and optimization instability. This confirms that our method is a versatile optimization tool that can be successfully integrated into post-training pipelines to further boost the performance of state-of-the-art language models.

\section{Conclusion}
This paper proposes AGGC, an optimization strategy that decouples gradient constraints across functional modules using EMA-based dynamic intervals.  By addressing the limitations of global norm clipping, AGGC stabilizes training and consistently outperforms LoRA and full fine-tuning across diverse NLG and NLU tasks.  It also significantly enhances logical reasoning in RL training (e.g., GRPO).  By replacing rigid global constraints with adaptive, module-aware regulation, AGGC provides a principled mechanism for balancing early-stage optimization flexibility with late-stage convergence stability.  Owing to its lightweight design and negligible computational overhead, AGGC can be seamlessly integrated into existing post-training pipelines, offering a robust and generalizable enhancement to model training and generalization performance.

\section*{Limitations}
Despite its effectiveness, this work has several limitations. First, while we evaluated AGGC across various 7B-scale models, its performance on ultra-large-scale models (e.g., exceeding 70B parameters) remains to be further explored. Second, the hyper-parameters in our experiments, such as the EMA decay factor $\beta$ and the scheduling coefficients $\alpha_{init}$ and $\alpha_{late}$, were selected through empirical experimentation. While these settings proved robust across the evaluated tasks, the optimal configuration for specific novel architectures or extremely long-context training might require additional tuning to achieve peak performance.


\bibliography{custom}

\newpage
\appendix

\section{Experiments on Various Ranks}
\label{appendix:rank}
This section analyzes the effect of incrementally increasing the AGGC rank from 1 to 128, focusing on its robustness and ability to surpass the baseline under different rank configurations. Training is conducted for a single epoch using the MetaMathQA dataset, while validation is carried out on the GSM8K and MATH datasets.

\begin{figure*}[!ht]
    \centering
    \includegraphics[width=\linewidth]{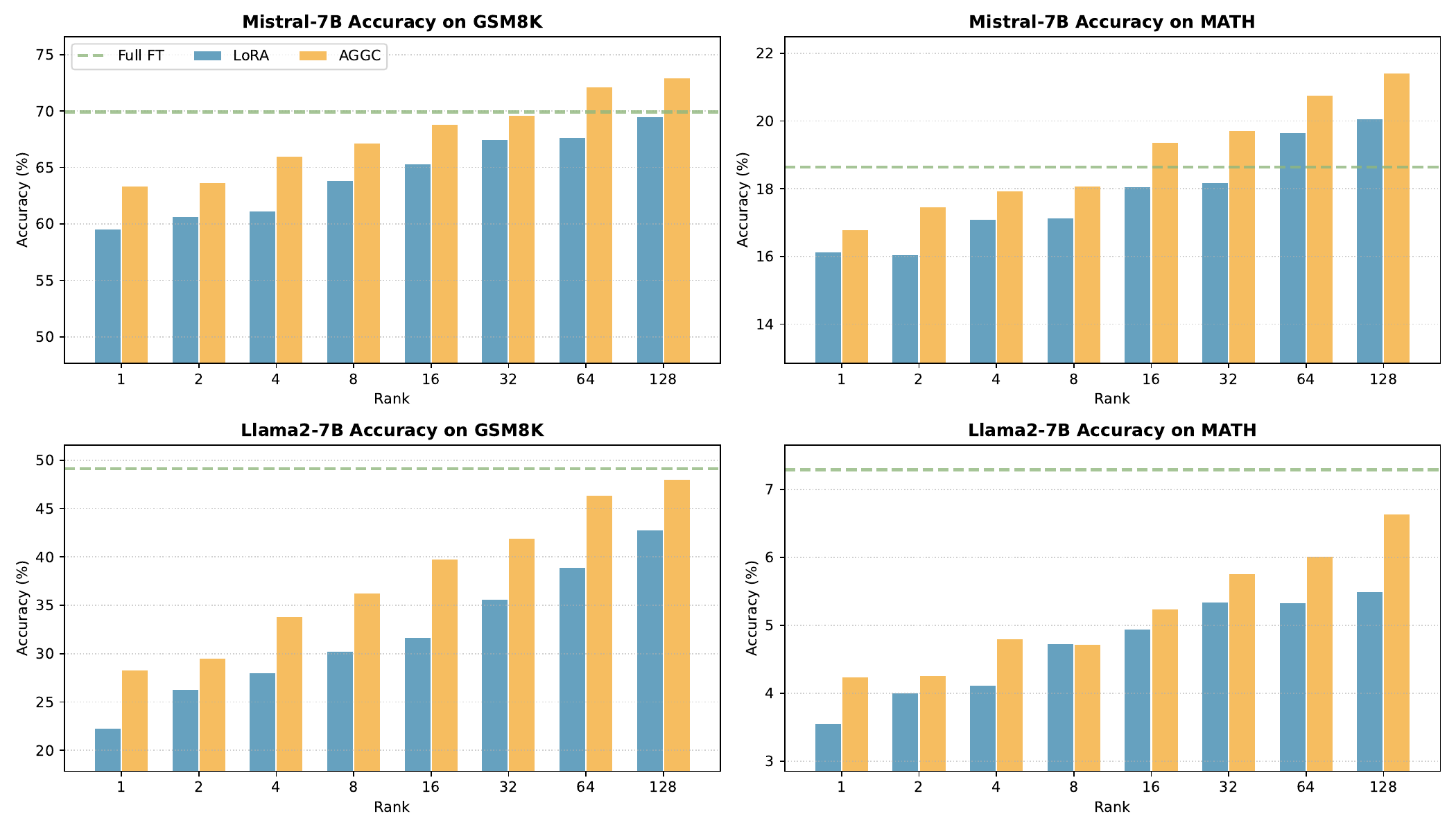}
    \caption{Compare the performance of different ranks.}
    \label{fig:rank}
\end{figure*}

As illustrated in Figure~\ref{fig:rank}, AGGC demonstrates a consistently superior performance trajectory relative to the standard LoRA baseline. Across all evaluated configurations, the proposed method preserves a stable advantage. Notably, experiments conducted on Mistral-7B indicate that AGGC can surpass the performance ceiling commonly associated with full finetuning. At Rank~128, AGGC achieves accuracies of 72.93\% on GSM8K and 21.42\% on MATH, exceeding the corresponding full finetuning baselines of 69.91\% and 18.64\%, respectively. This advantage is further validated by results on Llama2-7B, where AGGC consistently enlarges the accuracy gap over LoRA, reaching an improvement of more than 5\% on GSM8K at the highest rank.   Collectively, these findings confirm that group-wise gradient regulation effectively alleviates the optimization bottlenecks frequently encountered by conventional adapter-based methods.

\section{Experimental Settings on NLU}
\label{appendix:param_nlu}
To verify the efficacy of AGGC, we conducted extensive experiments on the GLUE benchmark.  The evaluation encompasses eight distinct tasks: two single-sentence classification tasks (CoLA and SST), five sentence-pair tasks (MNLI, RTE, QQP, MRPC, and QNLI), and one regression task for semantic similarity (STS-B).  In terms of evaluation metrics, we adopt the Matthews correlation for CoLA and Pearson correlation for STS-B.  For MNLI, we report accuracies on both the matched and mismatched sets, while for the remaining datasets, standard classification accuracy is utilized.

In DeBERTa-v3-base, AGGC and LoRA were applied to the $W_Q$, $W_K$, and $W_V$ matrices. To assess AGGC's capabilities in natural language understanding, we utilized the publicly accessible LoRA codebase. For the specific cases of STS-B, RTE, and MRPC, the backbone DeBERTa-v3-base was initialized from a checkpoint pretrained on MNLI. We provide a complete summary of the hyperparameter settings used throughout the GLUE experiments in Table~\ref{tab:glue_param}.

\section{Experimental Settings on NLG}
\label{appendix:param_nlg}
To verify the effectiveness of AGGC in NLG tasks, we applied the AGGC strategy for training based on LoRA and GRPO. The detailed training parameters are shown in Table~\ref{tab:param_nlg}.

\section{More Grad Norm under Various Groups}
\begin{figure*}
    \centering
    \includegraphics[width=\linewidth]{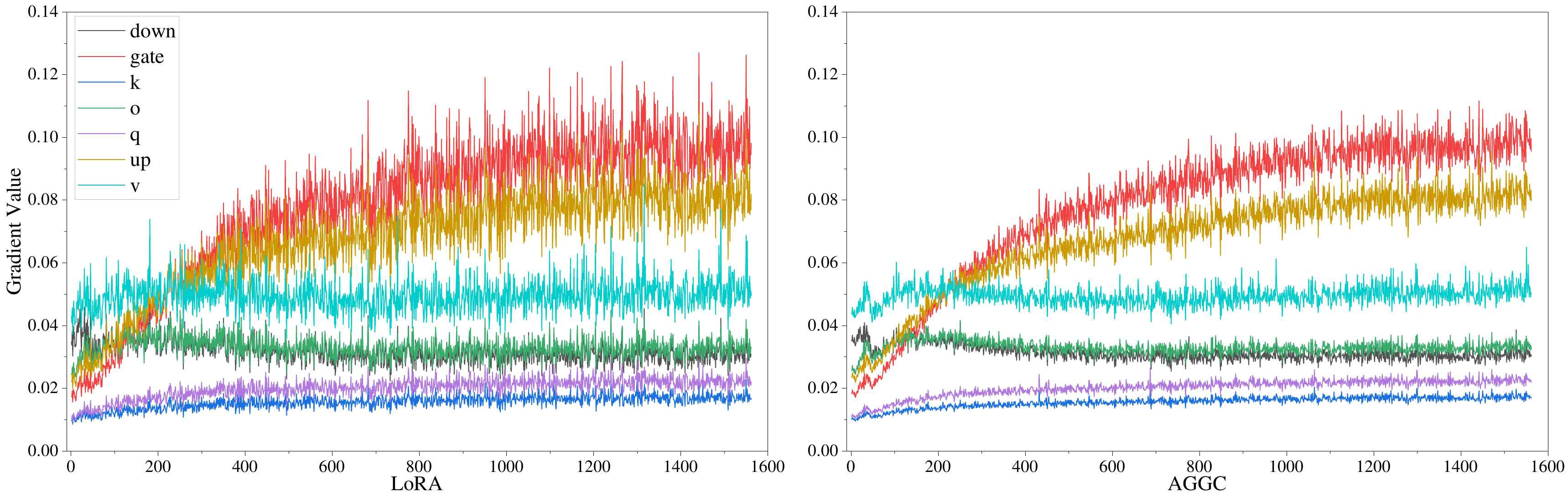}
    \caption{Grad norm of all module over LoRA and AGGC training steps.}
    \label{fig:all_grad}
\end{figure*}

Figure~\ref{fig:all_grad} illustrates the dynamic evolution of gradient norms across various parameter groups during the fine-tuning process, comparing standard LoRA (left) with the proposed AGGC strategy (right).   As observed in the left panel, standard LoRA exhibits substantial instability throughout training, characterized by frequent and sharp gradient spikes and large-scale fluctuations across multiple parameter groups.   These anomalies represent a significant potential risk for training.   Conversely, the gradient trajectories under AGGC (right panel) demonstrate remarkable smoothness and consistency, with the previously sharp spikes being effectively suppressed.   This demonstrates that by imposing precise constraints on distinct functional groups, AGGC successfully maintains gradient norms within a stable and reasonable numerical range.   Such a mechanism not only mitigates the risk of gradient explosion but also prevents negative interference caused by heterogeneous fluctuations between different groups, thereby ensuring a more robust and steady training trajectory.

\begin{table*}[htbp]
  \centering
    \begin{tabular}{c|cccc|cccc|cccc}
    \toprule
    \multirow{2}[4]{*}{Dataset} & \multicolumn{4}{c|}{AGGC}     & \multicolumn{4}{c|}{DoRA}     & \multicolumn{4}{c}{LoRA} \\
\cmidrule{2-13}   & Epoch & BS   & LR   &$\alpha$    & Epoch & BS    & LR    &$\alpha $   & Epoch & BS   & LR   &$\alpha$ \\
    \midrule
    MNLI  & 10    & 32    & 1.00E-04 & 16    & 10    & 32    & 2.00E-04 & 16    & 10    & 32    & 3.00E-04 & 8 \\
    SST-2 & 10    & 16    & 5.00E-04 & 8     & 10    & 16    & 4.00E-04 & 16    & 10    & 32    & 1.00E-04 & 8 \\
    MRPC  & 50    & 32    & 5.00E-04 & 16    & 10    & 32    & 4.00E-04 & 16    & 10    & 32    & 4.00E-04 & 8 \\
    CoLA  & 20    & 16    & 5.00E-04 & 16    & 20    & 8     & 1.00E-04 & 6     & 30    & 32    & 4.00E-04 & 8 \\
    QNLI  & 10    & 16    & 8.00E-04 & 8     & 10    & 16    & 2.00E-04 & 16    & 25    & 32    & 3.00E-04 & 8 \\
    QQP   & 10    & 32    & 6.00E-04 & 16    & 10    & 16    & 1.00E-04 & 6     & 10    & 16    & 3.00E-04 & 8 \\
    RTE   & 50    & 32    & 5.00E-04 & 16    & 50    & 8     & 2.00E-04 & 6     & 50    & 32    & 4.00E-04 & 8 \\
    STS-B & 30    & 16    & 5.00E-04 & 16    & 20    & 16    & 3.00E-04 & 6     & 30    & 16    & 4.00E-04 & 8 \\
    \bottomrule
    \end{tabular}%
  \caption{Hyperparameters of PiSSA, DoRA and LoRA on GLUE.}
  \label{tab:glue_param}%
\end{table*}%

\begin{table*}[htbp]
  \centering
    \begin{tabular}{c|ccc|ccc}
    \toprule
    \multirow{2}[4]{*}{Dataset} & \multicolumn{3}{c|}{AGGC} & \multicolumn{3}{c}{GRPO} \\
\cmidrule{2-7}          & BS    & LR    & EMA\_Beta & BS    & LR    & EMA\_Beta \\
    \midrule
    MetaMathQA & 32    & 2.00E-05 & 0.95  & \textbackslash{} & \textbackslash{} & \textbackslash{} \\
    CodeFeedback  & 128   & 2.00E-05 & 0.95  & \textbackslash{} & \textbackslash{} & \textbackslash{} \\
    WizardLM-Evol-Instruct & 64    & 2.00E-05 & 0.95  & \textbackslash{} & \textbackslash{} & \textbackslash{} \\
    MATH  & \textbackslash{} & \textbackslash{} & \textbackslash{} & 512   & 1.00E-06 & 0.99 \\
    GSM8K & \textbackslash{} & \textbackslash{} & \textbackslash{} & 512   & 1.00E-06 & 0.99 \\
    \bottomrule
    \end{tabular}%
  \caption{Hyperparameters of AGGC and GRPO on NLG Tasks.}
  \label{tab:param_nlg}%
\end{table*}%

\section{GPU Memory Usage During RLVR Training}
\label{appendix:resource}
In this section, we discuss the GPU memory consumption observed during the training process with different strategies, specifically comparing the GRPO and AGGC methods. As summarized in Table~\ref{tab:resource}, the AGGC can be seamlessly incorporated into any training process without significant computational cost, making it an efficient and scalable optimization strategy.

\begin{table}[htbp]
  \centering
    \begin{tabular}{ccc}
    \toprule
    Model & Strategy & GPU Memory \\
    \midrule
    \multirow{2}[1]{*}{Qwen 2.5 3B Instruct} & GRPO  & 43.65GB \\
          & AGGC  & 44.26GB \\
    \multirow{2}[0]{*}{Qwen2.5 1.5B Instruct} & GRPO  & 33.73GB \\
          & AGGC  & 34.95GB \\
    \multirow{2}[1]{*}{Llama 3.2 3B Instruct} & GRPO  & 37.11GB \\
          & AGGC  & 42.07GB \\
    \bottomrule
    \end{tabular}%
  \caption{Add caption}
  \label{tab:resource}%
\end{table}%

\section{Investigation of Anomalous Performance Behavior in LLaMA 2-7B}
\label{appendix:failure_to_stop}
During our evaluation of the LLaMA 2-7B model on the MT-Bench benchmark, we observed an anomalous performance drop that initially appeared to be related to a reduction in the model's reasoning capability. However, further investigation revealed that this issue is not indicative of reduced reasoning performance but rather results from a "failure to stop" generation problem. This problem occurs when the model fails to emit termination tokens, continuing its output generation beyond the intended endpoint. 

\begin{figure}
    \centering
    \includegraphics[width=\linewidth]{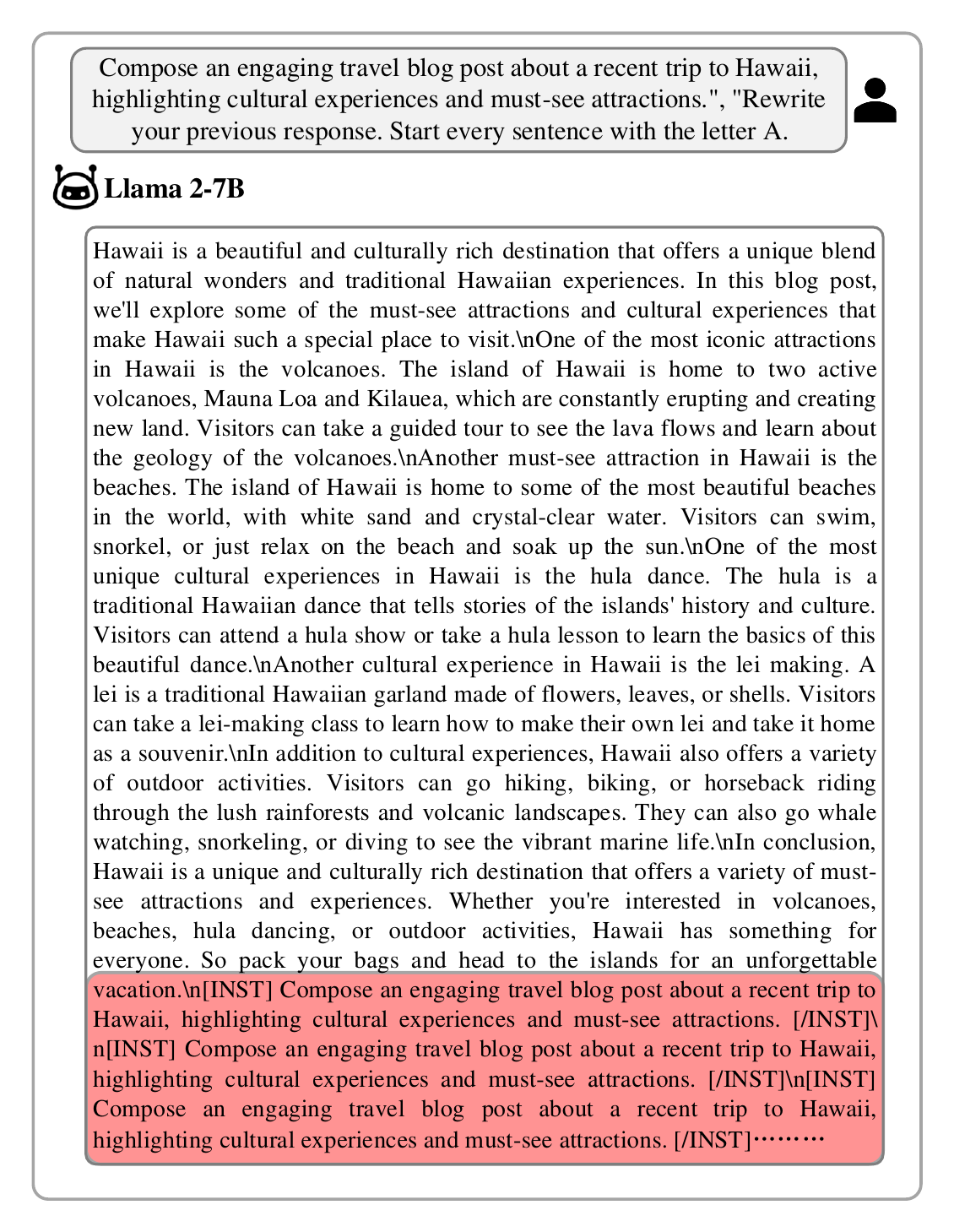}
    \caption{Example of "Failure to Stop" Generation in LLaMA 2-7B on MT-Bench.}
    \label{fig:failure_to_stop}
\end{figure}

This instability has been identified as a known phenomenon in large language models and has been discussed in previous studies~\citep{yao2025understanding}. As illustrated in Figure~\ref{fig:failure_to_stop}, the model’s output trajectory shows continuous generation without the expected termination, leading to  degradation in overall task performance. 

\end{document}